\documentclass[conference,a4paper]{IEEEtran}

\usepackage[utf8]{inputenc} 
\usepackage[T1]{fontenc}
\usepackage{url}              
\usepackage{cite}             

\usepackage[cmex10]{amsmath}  
\usepackage{mleftright}       
\mleftright                   

\usepackage{graphicx}         
\usepackage{booktabs}         

\usepackage[utf8]{inputenc} 
\usepackage[T1]{fontenc}    
\usepackage{url}            
\usepackage{booktabs}       
\usepackage{amsfonts}       
\usepackage{nicefrac}       
\usepackage{microtype}      
\usepackage{xcolor}         
\usepackage{amsmath, amssymb}      
\usepackage{enumerate}
\usepackage{cuted}
\usepackage{lipsum}
\usepackage{balance}

\usepackage{tikz} 
\usepackage{pgfplots}
\usepackage{pgfplotstable}
\usetikzlibrary{patterns}
\pgfplotsset{compat = newest}
\pgfkeys{/pgf/number format/set thousands separator = }
\usepackage{filecontents}
\usetikzlibrary{decorations.pathmorphing, graphs,graphs.standard} 
\usetikzlibrary{automata}
\usepgfplotslibrary{statistics}
\pgfplotsset{compat = newest}
\pgfkeys{/pgf/number format/set thousands separator = }
\definecolor{chocolate1}{rgb}{0.94,0.7,0.5}

\newcommand{\bigomega}{\makebox{\Large\ensuremath{\omega}}}

\newcommand{\R}{\mathbb{R}}
\newcommand{\C}{{\mathbb C}}

\newcommand{\calK}{\mathcal{K}}

\newcommand{\G}{\mathcal{G}}
\newcommand{\V}{\mathcal{V}}

\newcommand{\E}{\mathcal{E}}

\newcommand{\X}{\mathbf{X}}
\newcommand{\x}{{\mathbf x}}

\newcommand{\ccirc}{\raisebox{0.6pt}{\textcircled{\raisebox{-.3pt} {r}}}}

\newcounter{examplecntr}
{\begin{trivlist}\small\item[]\refstepcounter{examplecntr}%
 {\bfseries Example~\theexamplecntr%
  \ifthenelse{\equal{#1}{}}{}{ (#1)}.
}}%
{\end{trivlist}}

\newcounter{propositioncntr}
{\begin{trivlist}\item[]\refstepcounter{propositioncntr}%
{\bfseries Proposition~\thepropositioncntr%
  \ifthenelse{\equal{#1}{}}{}{ (#1)}.
}}%
{\end{trivlist}}

\newcounter{remarkcntr}
{\begin{trivlist}\item[]\refstepcounter{remarkcntr}%
{\bfseries Remark~\theremarkcntr%
  \ifthenelse{\equal{#1}{}}{}{ (#1)}.
}}%
{\end{trivlist}}

\newcounter{theoremcntr}
{\begin{trivlist}\item[]\refstepcounter{theoremcntr}%
{\bfseries Theorem~\thetheoremcntr%
  \ifthenelse{\equal{#1}{}}{}{ (#1)}.
}}%
{\hfill$\Box$\end{trivlist}}

\begin{document}

\title{The Exact Determinant of a\\  Specific Class of Sparse Positive Definite Matrices} 

\author{\IEEEauthorblockN{Mehdi Molkaraie}
\IEEEauthorblockA{Universitat Pompeu Fabra \\
08018 Barcelona, Spain \\
{\tt mehdi.molkaraie@alumni.ethz.ch}}
}

\maketitle

\begin{abstract}
For a specific class of sparse Gaussian graphical models, we provide a closed-form 
solution for the determinant of the covariance matrix. In our framework, 
the graphical interaction model (i.e., the covariance selection model) is equal to replacement
product of $\mathcal{K}_{n}$ and $\mathcal{K}_{n-1}$, where $\mathcal{K}_n$ is the complete graph with $n$ vertices. 
Our analysis is based on
taking the Fourier transform of the local factors of the model, which can be
viewed as an application of the Normal Factor Graph Duality Theorem and holographic algorithms. 
The closed-form expression is obtained by applying the Matrix Determinant Lemma on the 
transformed graphical model. 
In this context, we will also define a notion of equivalence between 
two Gaussian graphical models.
\end{abstract}

\section{INTRODUCTION}
\label{sec:intro}

A zero-mean real random vector ${\X}_{p\times 1}$ has a $p$-variate Gaussian distribution if it 
has the following PDF
\begin{equation}
\label{eqn:GaussPDF}
p(\x) =  \frac{1}{\text{det}(2\pi \pmb{\Sigma})^{1/2}}
\textrm{exp}\big(-\frac{1}{2}
\,\x^\intercal
\pmb{\Sigma}^{-1} 
\x\big), \quad \x \in \R^p
\end{equation}
where $\x = (x_1, x_2, \ldots, x_p)^\intercal$,
the symmetric positive definite matrix $\pmb{\Sigma}^{-1} \in \R^{p\times p}$ is the 
information (precision) matrix, and $\mathsf{\Sigma}$ is the corresponding covariance matrix.

The structure of a Gaussian graphical model is completely specified by its information matrix. A
nonzero entry of $\pmb{\Sigma}^{-1}$ indicates the presence of a factor in the graphical model, 
whereas an off-diagonal zero entry of $\pmb{\Sigma}^{-1}$ indicates the lack of pairwise interaction between
the corresponding random variables~\cite{lauritzen1996graphical}, \cite[Chapter 19]{murphy2012machine}. 
We use graphical models defined in terms of normal factor graphs (NFGs), in which variables are represented by edges and factors by vertices. Moreover, NFGs allow for a 
simple and elegant graph dualization procedure~\cite{Forney:01}. 

In this paper, we are concerned with computing $\text{det$(\pmb{\Sigma})$}$.
The determinant of a covariance/information matrix is an important quantity in many areas, e.g., in computing the generalized variance~\cite{wilks1932certain, Sengupta2004}, the differential 
entropy~\cite{ahmed1989entropy, misra2005estimation}, and the \mbox{Kullback–Leibler} divergence~\cite{cover1991elements}.
We assume that the information matrix and hence the corresponding NFG are sparse, in the sense that the number of non-zero entries 
of $\pmb{\Sigma}^{-1}$ is $\mathcal{O}(p)$, i.e., the number of non-zero entries of the information matrix is 
roughly equal to the number of its rows. 

In general, $\text{det$(\pmb{\Sigma})$}$ can be obtained via the Cholesky decomposition with computational complexity $\mathcal{O}(p^3)$, which is impractical 
when $p$ is large. Lower and upper bounds for the determinant of sparse positive definite matrices are provided in~\cite{bai1996bounds}.
Monte Carlo methods have been proposed in~\cite{bai1996some, barry1999monte} to estimate the determinant (or the log determinant) of 
sparse (positive definite) matrices. Randomized algorithms have been introduced to approximate the log determinant of positive definite 
matrices (see, e.g., \cite{boutsidis2017randomized}, where the computational complexity of the proposed algorithm depends on the 
number of nonzero elements of the matrix.). Further techniques to approximate the determinant of (sparse) positive definite matrices have 
been suggested, including sparse approximate inverses~\cite{reusken2002approximation} and Chebyshev polynomial expansions \cite{han2015large}.
However, in this paper, for a class of sparse Gaussian graphical models, we provide a \emph{closed-form} solution to $\text{det$(\pmb{\Sigma})$}$.

The paper is structured as follows. Some notation and preliminaries are introduced in Section~\ref{sec:notations}. The model 
and its NFG are discused in Section~\ref{sec:NFG}. In Section~\ref{sec:CovSel} the covariance selection model is described via the replacement product of graphs. The information 
matrix of the model is described in Section~\ref{sec:InfMatrix}. The closed-form solution of the determinant is 
derived in Section~\ref{sec:DUALNFG}. Numerical examples are presented in 
Section \ref{sec:ExampleT}.

\section{Notation and preliminaries}
\label{sec:notations}

In this section, we introduce notation and preliminaries that will be used throughout this paper.

The all-ones vector with size $p\times 1$ is denoted by $\mathbf{1}_{p}$, the all-ones matrix with 
size $p\times p$ is denoted by $\mathbf{J}_{p}$, and the identity matrix with size 
$p\times p$ is denoted by $\mathbf{I}_{p}$. The trace of a matrix $\mathbf{B}$ is denoted by $\text{tr$(\mathbf{B})$}$.

$\G = (\V, \E)$ is a graph with vertex set $\V$ and edge set $\E$. We 
assume that $\G$ is simple, connected, and undirected.
A graph $\G$ is called \emph{regular} if every vertex has the same degree (i.e., if each vertex has the same number of neighbours).
The complete graph (i.e., the fully connected graph) with $n$ vertices is denoted by $\calK_n$ which is a regular 
graph of degree $n-1$. The number of edges of $\calK_n$ is $|\E| = n(n-1)/2$. Also let 
\begin{equation}
\label{eqn:DefN}
N = n(n-1)
\end{equation}
which is equal to $2|\E|$. The replacement product of two graphs $\G_1$ and $\G_2$ is denoted 
by $\G_1 \ccirc \G_2$, see~\cite{reingold2000entropy, alon2008elementary}.

The set of positive integers from 1 to $n$ is denoted by $[n]$. We sometimes refer to the vertex $V_i$
of $\calK_n$ by its index $i$ which takes values in $[n]$. For a subset $\mathcal{T} \subset [n]$, let $\x_\mathcal{T} = (x_t, t \in \mathcal{T})$ and define the zero-sum indicator function as
\begin{equation} 
\label{eqn:Definition2}
\delta_{+}(\x_\mathcal{T}) =  \left\{ \begin{array}{ll}
    1, & \text{if $x_1 + x_2 + \ldots + x_{|\mathcal{T}|} = 0$} \\
    0, & \text{otherwise,}
  \end{array} \right.
\end{equation}
and the equality indicator function as
\begin{equation} 
\label{eqn:Definition1}
\delta_{=}(\x_\mathcal{T}) =  \left\{ \begin{array}{ll}
    1, & \text{if $x_1 = x_2 = \ldots = x_{|\mathcal{T}|}$} \\
    0, & \text{otherwise.}
  \end{array} \right.
\end{equation}
Notice that for $|\mathcal{T}|=1$, both (\ref{eqn:Definition2}) and (\ref{eqn:Definition1}) are equivalent to the dirac 
delta function, denoted by $\delta(\cdot)$. 

Our analysis relies on the Fourier transform of the local factors of the NFG.
The Fourier transform of a function $f(\x) \colon \R^p \rightarrow \C$ is the function
$(\mathcal{F}f)(\pmb{\omega}) \colon \R^p \rightarrow \C$ given by
\begin{equation}
\label{eqn:ft}
(\mathcal{F}f)(\pmb{\omega}) = \int_{-\infty}^{\infty} f(\x)e^{-\mathrm{i}\,\pmb{\omega}^\intercal \x}d\x
\end{equation}
where $\pmb{\omega} = (\omega_1, \omega_2, \ldots, \omega_p)^\intercal$, $\mathrm{i} = \sqrt{-1}$, and $\C$ denotes the set of 
complex numbers~\cite{osgood2019lectures}. In particular, the Fourier transform of the PDF in~(\ref{eqn:GaussPDF}) is
\begin{equation}
\label{eqn:ftGaussian}
(\mathcal{F}p)(\pmb{\omega}) = \textrm{exp}\big(-\frac{1}{2}
\,\pmb{\omega}^\intercal
\pmb{\Sigma} 
\,\pmb{\omega}\big), \quad \pmb{\omega} \in \R^p
\end{equation}
see~\cite{anderson1958introduction}, and the Fourier transform of (\ref{eqn:Definition2}) can be computed as
\begin{align}
(\mathcal{F}\delta_{+})(\pmb{\omega}) & = \int_{-\infty}^{\infty} \delta_{+}(\x)e^{-\mathrm{i}\,\pmb{\omega}^\intercal \x}d\x \\
& = \int_{-\infty}^{\infty} \delta(x_1 + x_2 + \ldots + x_p)e^{-\mathrm{i}\,\pmb{\omega}^\intercal \x}d\x \\
& = (2\pi)^{p-1}\prod_{1\le i < p} \delta(\omega_i - \omega_p)
\end{align}
Hence
\begin{equation}
\label{eqn:ftdeltaplus}
(\mathcal{F}\delta_{+})(\pmb{\omega}) = (2\pi)^{p-1}\delta_{=}(\pmb{\omega})
\end{equation}

In other words $\delta_{+}(\cdot)$ and $\delta_{=}(\cdot)$ are Fourier transform pairs up to a scale factor.


\section{The model and the NFG}
\label{sec:NFG}

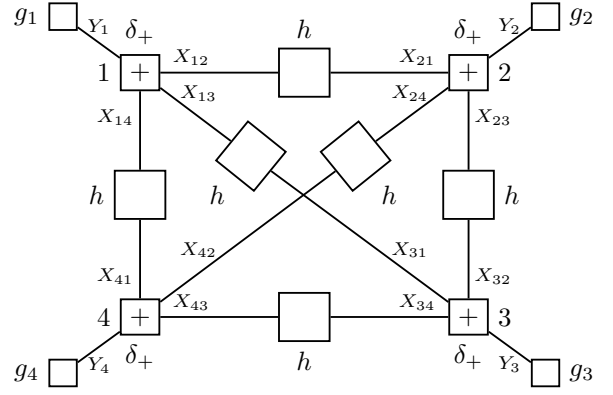
\begin{figure}[t]
  \centering
  \begin{tikzpicture}[scale=0.27]
   \linethickness{1.7mm}
%
%
%
%
%

\node[draw, line width=0.7, rectangle, label=below:{$\delta_{+}$}, label=left:{$4$}] at (0,0) (0) {$+$};
\node[draw, line width=0.7, rectangle, label=below:{$h$}] at (8,0) (1) {\textcolor{white}{{\LARGE A}}};
\node[draw, line width=0.7, rectangle, label=below:{$\delta_{+}$}, label=right:{$3$}] at (16,0) (2) {$+$};
\node[draw, line width=0.7, rectangle, label=left:{$h$}] at (0,6) (3) {\textcolor{white}{{\LARGE A}}};
\node[draw, line width=0.7, rectangle, label=above:{$\delta_{+}$},  label=left:{$1$}] at (0,12) (4) {$+$};
\node[draw, line width=0.7, rectangle, label=above:{$h$}] at (8,12) (5) {\textcolor{white}{{\LARGE A}}};
\node[draw, line width=0.7, rectangle, label=above:{$\delta_{+}$}, label=right:{$2$}] at (16,12) (6) {$+$};
\node[draw, line width=0.7, rectangle, label=right:{$h$}] at (16,6) (7) {\textcolor{white}{{\LARGE A}}};
\node[draw, line width=0.7, rotate=-51, rectangle, label=right:{$h$}] at (10.5,7.9) (8) {\textcolor{white}{{\LARGE A}}};
\node[draw, line width=0.7, rotate=51, rectangle, label=left:{$h$}] at (5.4,7.9) (9) {\textcolor{white}{{\LARGE A}}};
\node[draw, line width=0.7, rectangle, label=left:{$g_{4}$}] at (-3.75,-2.75) (10) {\textcolor{white}{{\tiny A}}};
\node[draw, line width=0.7, rectangle, label=right:{$g_{3}$}] at (19.75,-2.75) (11) {\textcolor{white}{{\tiny A}}};
\node[draw, line width=0.7, rectangle, label=right:{$g_{2}$}] at (19.75,14.75) (12) {\textcolor{white}{{\tiny A}}};
\node[draw, line width=0.7, rectangle, label=left:{$g_{1}$}] at (-3.75,14.75) (13) {\textcolor{white}{{\tiny A}}};
\draw[line width=0.7] (0) -- (1);
\draw[line width=0.7] (1) -- (2);
\draw[line width=0.7] (0) -- (3);
\draw[line width=0.7] (3) -- (4);
\draw[line width=0.7] (4) -- (5);
\draw[line width=0.7] (5) -- (6);
\draw[line width=0.7] (6) -- (7);
\draw[line width=0.7] (7) -- (2);
\draw[line width=0.7] (0) -- (8);
\draw[line width=0.7] (8) -- (6);
\draw[line width=0.7] (4) -- (9);
\draw[line width=0.7] (9) -- (2);
\draw[line width=0.7] (2) -- (11);
\draw[line width=0.7] (0) -- (10);
\draw[line width=0.7] (4) -- (13);
\draw[line width=0.7] (6) -- (12);
\draw (-2.0,14.25) node{${\scriptstyle Y_1}$};
\draw (18,14.28) node{${\scriptstyle Y_2}$};
\draw (18,-2.32) node{${\scriptstyle Y_3}$};
\draw (-2.0,-2.38) node{${\scriptstyle Y_4}$};
\draw (2.5,12.75) node{${\scriptstyle X_{12}}$};
\draw (2.85,11) node{${\scriptstyle X_{13}}$};
\draw (-1.2,9.8) node{${\scriptstyle X_{14}}$};
\draw (13.5,12.75) node{${\scriptstyle X_{21}}$};
\draw (13.15,11) node{${\scriptstyle X_{24}}$};
\draw (17.2,9.8) node{${\scriptstyle X_{23}}$};
\draw (2.5,0.73) node{${\scriptstyle X_{43}}$};
\draw (2.85,3.3) node{${\scriptstyle X_{42}}$};
\draw (-1.2,2) node{${\scriptstyle X_{41}}$};
\draw (17.2,2) node{${\scriptstyle X_{32}}$};
\draw (13.5,0.73) node{${\scriptstyle X_{34}}$};
\draw (13.15,3.3) node{${\scriptstyle X_{31}}$};
%
%
  \end{tikzpicture}
  \vspace{0.5ex}
  \caption{\label{fig:Primalfactors}
The NFG for the Gaussian graphical model defined on $\calK_4$ with PDF as in~(\ref{eqn:ModelPDF2}). 
The boxes labelled ``$+$" are zero-sum indicator factors (\ref{eqn:Definition2}) which enforce the constraint in~(\ref{eqn:defy}). 
The small empty boxes represent unary factors $\{g_i(y_i), i \in [n]\}$ given by~(\ref{eqn:defunaryg}) and the big empty 
boxes represent pairwise factors $\{h(x_{ij}, x_{ji}), (i,j) \in \E\}$ as in (\ref{eqn:defpairwiseh}).}
\end{figure}

We consider Gaussian graphical models defined on $\calK_n$. For $n=4$, the NFG of the model is illustrated in Fig.~\ref{fig:Primalfactors}, in
which $|\V| = n$ zero-sum indicator factors sit at the vertices, $|\V|$ unary factors are attached to 
the zero-sum indicator factors, and $|\E|$ pairwise factors are placed on the edges of the NFG.

We refer to the $i$-th zero-sum indicator factor by its index $i \in [n]$. Accordingly, let
\begin{equation}
\label{eqn:defxi}
\x_{\text{-}i} = \left(x_{ij}\right)_{j \in \left[n\right] \setminus \{i\}} \quad \x_{\text{-}i} \in \R^{n-1}
\end{equation}
denote the variables attached to the $i$-th zero-sum indicator factor, e.g., $\x_{\text{-}2} = (x_{21}, x_{23}, x_{24})$ in Fig.~\ref{fig:Primalfactors}. 
Regarding the labelling of the edges (i.e., the variables) of the NFG, we adopt the following convention: the pairwise 
factor placed on $(i,j) \in \E$ is only
a function of $X_{ij}$ and $X_{ji}$.



The $i$-th zero-sum  indicator factor $\delta_{+}(\x_{\text{-}i}, y_i)$ imposes the constraint that 
\begin{equation}
\label{eqn:defy}
y_i \;\, + \!\!\sum_{j \in \left[n\right] \setminus \{i\}} x_{ij} = 0
\end{equation}

Since $y_i$ is fully specified from $\x_{\text{-}i}$, the total number of independent variables in the 
model is equal to $N$ given by (\ref{eqn:DefN}). Indeed 
\begin{equation}
\x = (\x_{\text{-}1}, \x_{\text{-}2}, \x_{\text{-}3}, \ldots, \x_{\text{-}n})^{\intercal}_{} \quad \x \in \R^N
\end{equation}

For $i \in [n]$, we assume that the unary factor $g_i(\cdot)$ is a univariate Gaussian (up to scale) given by
\begin{equation}
\label{eqn:defunaryg}
g_i(y_i) = \textrm{exp}\big(-\frac{1}{2s_i^2}y_i^2 \big)
\end{equation}
where $s_i$ is the corresponding standard deviation.

For $(i, j) \in \E$ the pairwise factor $h(\cdot)$ is a bivariate Gaussian (up to scale) as in
\begin{equation}
\label{eqn:defpairwiseh}
h(x_{ij}, x_{ji}) = \textrm{exp}\bigg(-\frac{1}{2} 
\setlength\arraycolsep{2pt}
\begin{bmatrix}
x_{ij} & x_{ji}
\end{bmatrix}
\mathbf{K}^{-1}_{}
\begin{bmatrix}
x_{ij} \\
x_{ji}
\end{bmatrix}
\bigg)
\end{equation}
where the symmetric positive definite matrix $\mathbf{K}^{-1}_{}$ is given by
\begin{equation}
\label{eqn:K}
\mathbf{K}^{-1}_{}  = \frac{1}{1-\rho^2}
\left[ \begin{array}{cc}
  1/\sigma^2 & -\rho/\sigma^2  \\[0.4ex]
  -\rho/\sigma^2 & 1/\sigma^2
\end{array} \right]
\end{equation}
Here $\rho \in (-1,1)$ denotes a correlation coefficient. 

According to the adopted labelling of the vertices and the edges of the NFG (as illustrated in Fig.~\ref{fig:Primalfactors}), the PDF associated with the model is 
\begin{equation}
f(\x)  = \frac{1}{Z_\x}\prod_{i \in \left[n\right]} \delta_{+}(\x_{\text{-}i}, y_i)\prod_{i \in \left[n\right]} g_i(y_i) \prod_{(i, j) \in \E} h(x_{ij}, x_{ji}) \label{eqn:ModelPDF1}
\end{equation}

After substituting the constraint imposed by the zero-sum indicator factors in (\ref{eqn:defy}), we can express the PDF 
in (\ref{eqn:ModelPDF1}) as
\begin{IEEEeqnarray}{lCl}
f(\x) = \frac{1}{Z_\x}\!\!
 &\prod_{i \in \left[n\right]}&\textrm{exp}\big(-\frac{1}{2} \x^{\intercal}_{\text{-}i}\mathbf{S}^{}_{i}\x^{}_{\text{-}i}\big)\notag  \\ 
&\prod_{(i, j) \in \E}& \textrm{exp}\bigg(-\frac{1}{2} 
\setlength\arraycolsep{2pt}
\begin{bmatrix}
x_{ij} & x_{ji}
\end{bmatrix}
\mathbf{K}^{-1}_{}
\begin{bmatrix}
x_{ij} \\
x_{ji}
\end{bmatrix}
\bigg) \label{eqn:ModelPDF2}
\end{IEEEeqnarray}
where $Z_\x$ is the appropriate normalization constant and  
\begin{equation}
\label{eqn:S}
\mathbf{S}_i = \frac{1}{s_{i}^2}\mathbf{J}_{n-1}
\end{equation}
I.e., every entry of $\mathbf{S}_i$ is equal to $1/s_{i}^2$. The model is called \emph{homogeneous} if $s_i = s$ for $i \in [n]$.

Next, we will show how the graphical interaction model associated with Fig.~\ref{fig:Primalfactors} can be viewed as
the replacement product of two complete graphs.

\section{The covariance selection model via\linebreak the replacement product}
\label{sec:CovSel}

\begin{figure}[t]
  \centering
  \begin{tikzpicture}[scale=0.22]
   \linethickness{1.7mm}
%
%
%
%
%

\node[draw, line width=0.7, circle, fill=black, minimum size=1mm, label=below:{$X_{43}$}] at (0,0) (0) {};
\node[draw, line width=0.7, circle, fill=black, minimum size=1mm, label=below:{$X_{34}$}] at (16,0) (1) {$$};
\node[draw, line width=0.7, circle, fill=black, minimum size=1mm, label=above:{$X_{12}$}] at (0,12) (2) {$$};
\node[draw, line width=0.7, circle, fill=black, minimum size=1mm, label=above:{$X_{21}$}] at (16,12) (3) {$$};
\node[draw, line width=0.7, circle, fill=black, minimum size=1mm, label=right:{$X_{42}$}] at (3,3) (4) {$$};
\node[draw, line width=0.7, circle, fill=black, minimum size=1mm, label=left:{$X_{41}$}] at (-3,3) (5) {$$};
\node[draw, line width=0.7, circle, fill=black, minimum size=1mm, label=right:{$X_{32}$}] at (19,3) (6) {$$};
\node[draw, line width=0.7, circle, fill=black, minimum size=1mm, label=left:{$X_{31}$}] at (13,3) (7) {$$};
\node[draw, line width=0.7, circle, fill=black, minimum size=1mm, label=left:{$X_{14}$}] at (-3,9) (8) {$$};
\node[draw, line width=0.7, circle, fill=black, minimum size=1mm, label=right:{$X_{13}$}] at (3,9) (9) {$$};
\node[draw, line width=0.7, circle, fill=black, minimum size=1mm, label=left:{$X_{24}$}] at (13,9) (10) {$$};
\node[draw, line width=0.7, circle, fill=black, minimum size=1mm, label=right:{$X_{23}$}] at (19,9) (11) {$$};
\draw[line width=0.7] (0) -- (4);
\draw[line width=0.7] (0) -- (5);
\draw[line width=0.7] (4) -- (5);
\draw[line width=0.7] (1) -- (6);
\draw[line width=0.7] (1) -- (7);
\draw[line width=0.7] (6) -- (7);
\draw[line width=0.7] (8) -- (2);
\draw[line width=0.7] (9) -- (2);
\draw[line width=0.7] (8) -- (9);
\draw[line width=0.7] (9) -- (7);
\draw[line width=0.7] (4) -- (10);
\draw[line width=0.7] (3) -- (10);
\draw[line width=0.7] (3) -- (11);
\draw[line width=0.7] (10) -- (11);
\path [line width=0.7]
    (0) edge [bend right] node [below] {$$} (1)
    (2) edge [bend left] node [below] {$$} (3)
    (8) edge [bend right] node [below] {$$} (5)
    (6) edge [bend right] node [below] {$$} (11);

  \end{tikzpicture}
  \vspace{0.5ex}
  \caption{\label{fig:CovSel}
The covariance selection model of $\calK_{4} \ccirc\, \calK_{3}$ associated with Fig.~\ref{fig:Primalfactors}.}
\end{figure}
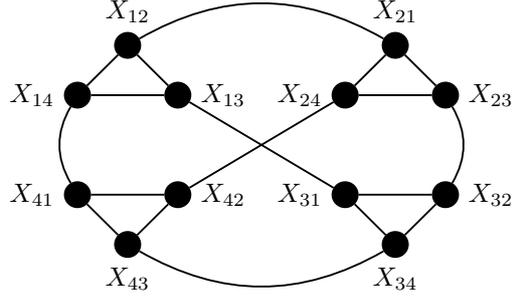

For $n=4$, the graphical interaction model (i.e., the covariance selection model~\cite{dempster1972covariance}) 
associated with Fig.~\ref{fig:Primalfactors} is illustrated in Fig.~\ref{fig:CovSel}, which is 
equal to $\calK_{4} \ccirc\, \calK_{3}$. 

Roughly speaking, the replacement product $\G = \G_1 \ccirc\, \G_2$ of two regular graphs $\G_1$ and $\G_2$ is obtained as follows.
Let $\G_1 = (\V_1, \E_1)$ and $\G_2 = (\V_2, \E_2)$ be two regular graphs of degree $d_1$ and $d_2$, respectively. Furthermore,
assume that $|\V_2| = d_1$. 
\begin{enumerate}
\item Replace each vertex of $\G_1$ with a copy (or \emph{cloud}) of $\G_2$, while maintaining the edges of $\G_1$ and $\G_2$.
\item Connect each vertex to all its neighbours inside its copy (according to $\G_2$) and to one vertex in a neighbouring 
copy (according to $\G_1$).
\end{enumerate}
The replacement product $\G = (\V, \E)$ is a $d_2+1$ regular graph with $|\V| = d_1|\V_1|$. For more details on the 
replacement product (and the zig-zag product) of graphs, see~\cite{reingold2000entropy}.

In our framework, the covariance selection model is equal to
$\calK_{n} \ccirc\, \calK_{n-1}$. Notice that the replacement product $\calK_{n} \ccirc\, \calK_{n-1}$ has exactly $N$ vertices -- each with degree $n-1$.


\section{The information matrix}
\label{sec:InfMatrix}

It is straightforward to show that the information matrix $\pmb{\Sigma}^{-1}_{\x}$ associated with $f(\x)$ in (\ref{eqn:ModelPDF2}) can be expressed as 
\begin{equation}
\label{eqn:IMPrimal}
\pmb{\Sigma}^{-1}_{\x}  = \frac{1}{(1-\rho^2)\sigma^2}\mathbf{I}_N + \mathbf{A}
\end{equation}
where $\mathbf{A} \in \R^{N\times N}$ has the following block structure
\begin{equation}
\label{eqn:Amatrix}
\mathbf{A}  = 
\left[ \begin{array}{ccccc}
  \mathbf{S}_{1} & \mathbf{E}_{1,2} & \mathbf{E}_{1,3} & \cdots &\mathbf{E}_{1,n} \\[0.5ex]
  \mathbf{E}^{\intercal}_{1,2} & \mathbf{S}_{2} & \mathbf{E}_{2,3} & \cdots &\mathbf{E}_{2,n} \\[0.5ex]
    \mathbf{E}^{\intercal}_{1,3} & \mathbf{E}^{\intercal}_{2,3} & \mathbf{S}_{3} & \cdots &\mathbf{E}_{3, n} \\[0.5ex]
  \vdots    &\vdots      & \vdots          & \ddots &\vdots    \\[0.5ex]
  \mathbf{E}^{\intercal}_{1,n} & \mathbf{E}^{\intercal}_{2,n} & \mathbf{E}^{\intercal}_{3, n} &\cdots &\mathbf{S}_{n}
\end{array} \right]
\end{equation}

Here $\mathbf{S}_i$ is as in (\ref{eqn:S}). For $1 \le i < j \le n$ the entry in the $k$-th row and the $m$-th column 
of $\mathbf{E}_{i, j}$ is given by
\begin{equation} \label{eqn:K2}
\mathbf{E}_{i,j}[k, m] = \left\{ \begin{array}{ll}
       \dfrac{-\rho}{(1-\rho^2)\sigma^2}, & \text{if $k = j-1$ and $m = i$,} \\[1.8ex]
       0, & \text{otherwise.} 
      \end{array}\right.   
\end{equation}
We emphasize that $\mathbf{E}_{i, j}$ is a single-entry matrix with only one nonzero 
entry at position $\mathbf{E}_{i,j}[\,j-1, i]$. 

All the submatrices in $\mathbf{A}$ are square matrices with $n-1$ rows, and therefore the symmetric positive definite information 
matrix $\pmb{\Sigma}^{-1}_{\x}$ has $N$ rows. Furthermore
$\pmb{\Sigma}^{-1}_{\x}$ is sparse, as each row of $\pmb{\Sigma}^{-1}_{\x}$ contains exactly $n$ nonzero entries.  

The normalization constant in~(\ref{eqn:ModelPDF2}) can be expressed as 
\begin{align}
Z_\x & = \text{det}(2\pi \pmb{\Sigma}_{\x})^{1/2} \\[0.25ex]
& = (2\pi)^{N/2}\text{det}(\pmb{\Sigma}_{\x})^{1/2} \label{eqn:Zx2}
\end{align}

Arbitrary relabellings (reorderings) of the edges of the NFG give rise to the following transformed information matrix
\begin{equation}
\label{eqn:transPerm}
\pmb{\Sigma}^{-1}_{\x} \mapsto\, \mathbf{P}\pmb{\Sigma}^{-1}_{\x}\mathbf{P}_{\textcolor{white}{a}}^{\intercal}
\end{equation}
 
Here $\mathbf{P}$ denotes the corresponding permutation matrix of conformable \mbox{size~\cite[Chapter 3]{saad2003iterative}}. 
The number of such permutation matrices is indeed $N!$. Clearly, the mapping in~(\ref{eqn:transPerm}) 
does not change the value of the normalization constant $Z_\x$. Our analytical 
solution can therefore be applied to a class of information matrices given by~(\ref{eqn:transPerm}) with 
$\pmb{\Sigma}^{-1}_{\x}$ as in (\ref{eqn:IMPrimal}).

\subsection{Example}
\label{sec:Example}

We consider $\calK_4$ with $\rho = 1/2$, $\sigma^2 = 2/3$, and $s^2_{i} = 1/i$ for $1 \le i \le 4$. 
According to (\ref{eqn:IMPrimal}), the $12\times 12$ information matrix is given by
\begin{equation}
\pmb{\Sigma}^{-1}_{\x}  = 2\mathbf{I}_{12} + \mathbf{A}
\end{equation}
where $\mathbf{A}$ is as in (\ref{eqn:Amatrix}), $\mathbf{S}_i = i\mathbf{J}_{3}$ for $1 \le i \le 4$, and for $1 \le i < j \le 4$ the nonzero entry of $\mathbf{E}_{i,j}$ 
is equal to $-1$. Thus 
\begin{multline*}
\pmb{\Sigma}^{-1}_{\x}  = \\ 
\resizebox{0.49\textwidth}{!}{$
\left[ \begin{array}{rrrrrrrrrrrr}
3 & 1 & 1 & -1 & 0 & 0 & 0 & 0 & 0 & 0 & 0 & 0\\
1 & 3 & 1 & 0 & 0 & 0 & -1 & 0 & 0 & 0 & 0 & 0\\
1 & 1 & 3 & 0 & 0 & 0 & 0 & 0 & 0 & -1 & 0 & 0\\
-1 & 0 & 0 & 4 & 2 & 2 & 0 & 0 & 0 & 0 & 0 & 0\\
0 & 0 & 0 & 2 & 4 & 2 & 0 & -1 & 0 & 0 & 0 & 0\\
0 & 0 & 0 & 2 & 2 & 4 & 0 & 0 & 0 & 0 & -1 & 0\\
0 & -1 & 0 & 0 & 0 & 0 & 5 & 3 & 3 & 0 & 0 & 0\\
0 & 0 & 0 & 0 & -1 & 0 & 3 & 5 & 3 & 0 & 0 & 0\\
0 & 0 & 0 & 0 & 0 & 0 & 3 & 3 & 5 & 0 & 0 & -1\\
0 & 0 & -1 & 0 & 0 & 0 & 0 & 0 & 0 & 6 & 4 & 4\\
0 & 0 & 0 & 0 & 0 & -1 & 0 & 0 & 0 & 4 & 6 & 4\\
0 & 0 & 0 & 0 & 0 & 0 & 0 & 0 & -1 & 4 & 4 & 6
\end{array} \right]$}
\end{multline*}


\vspace{1mm}

\section{Holographic algorithms and the dual NFG}
\label{sec:DUALNFG}

Following~\cite{Forney:01, Forney:11, Forney:18}, we can construct the dual of an NFG
by adopting the following steps: I) replace each variable, say $x_i$, by the dual variable $\omega_i$, II) replace 
each factor, say $g_i(\cdot)$, by its Fourier transform $(\mathcal{F}g_i)(\cdot)$, and III) replace each edge by a 
sign-inverting edge (by placing a zero-sum indicator factor in the middle of each edge). 

According to the NFG Duality Theorem, 
the partition function of an NFG and the partition function of its dual are equal up to scale~\cite{AY:2011}.
The NFG Duality Theorem can be viewed as the Holographic transformation, where
an NFG is transformed by changing all the local factors (e.g., via the Fourier transform).
For more
details (and connections to planar matching polynomials), see~\cite{valiant2008holographic, cai2008holographic, AY:2011}. 

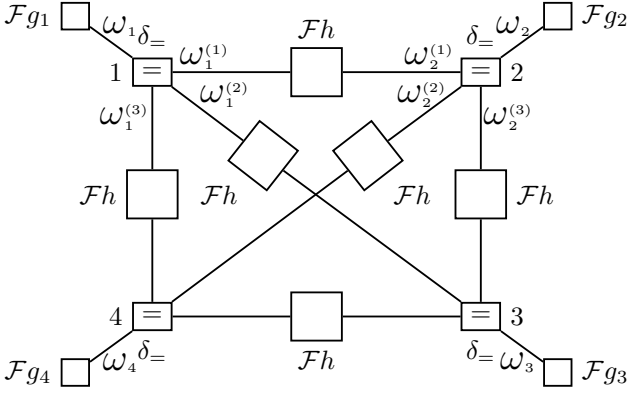
\begin{figure}[t]
  \centering
  \begin{tikzpicture}[scale=0.27]
   \linethickness{1.7mm}
%
%
%
%
%

\node[draw, line width=0.7, rectangle, label=below:{$\delta_{=}$}, label=left:{$4$}] at (0,0) (0) {$=$};
\node[draw, line width=0.7, rectangle, label=below:{$\mathcal{F}h$}] at (8,0) (1) {\textcolor{white}{{\LARGE A}}};
\node[draw, line width=0.7, rectangle, label=below:{$\delta_{=}$}, label=right:{$3$}] at (16,0) (2) {$=$};
\node[draw, line width=0.7, rectangle, label=left:{$\mathcal{F}h$}] at (0,6) (3) {\textcolor{white}{{\LARGE A}}};
\node[draw, line width=0.7, rectangle, label=above:{$\delta_{=}$},  label=left:{$1$}] at (0,12) (4) {$=$};
\node[draw, line width=0.7, rectangle, label=above:{$\mathcal{F}h$}] at (8,12) (5) {\textcolor{white}{{\LARGE A}}};
\node[draw, line width=0.7, rectangle, label=above:{$\delta_{=}$}, label=right:{$2$}] at (16,12) (6) {$=$};
\node[draw, line width=0.7, rectangle, label=right:{$\mathcal{F}h$}] at (16,6) (7) {\textcolor{white}{{\LARGE A}}};
\node[draw, line width=0.7, rotate=-52, rectangle, label=right:{$\mathcal{F}h$}] at (10.5,7.9) (8) {\textcolor{white}{{\LARGE A}}};
\node[draw, line width=0.7, rotate=52, rectangle, label=left:{$\mathcal{F}h$}] at (5.4,7.9) (9) {\textcolor{white}{{\LARGE A}}};
\node[draw, line width=0.7, rectangle, label=left:{$\mathcal{F}g_{4}$}] at (-3.75,-2.75) (10) {\textcolor{white}{{\tiny A}}};
\node[draw, line width=0.7, rectangle, label=right:{$\mathcal{F}g_{3}$}] at (19.75,-2.75) (11) {\textcolor{white}{{\tiny A}}};
\node[draw, line width=0.7, rectangle, label=right:{$\mathcal{F}g_{2}$}] at (19.75,14.75) (12) {\textcolor{white}{{\tiny A}}};
\node[draw, line width=0.7, rectangle, label=left:{$\mathcal{F}g_{1}$}] at (-3.75,14.75) (13) {\textcolor{white}{{\tiny A}}};
\draw[line width=0.7] (0) -- (1);
\draw[line width=0.7] (1) -- (2);
\draw[line width=0.7] (0) -- (3);
\draw[line width=0.7] (3) -- (4);
\draw[line width=0.7] (4) -- (5);
\draw[line width=0.7] (5) -- (6);
\draw[line width=0.7] (6) -- (7);
\draw[line width=0.7] (7) -- (2);
\draw[line width=0.7] (0) -- (8);
\draw[line width=0.7] (8) -- (6);
\draw[line width=0.7] (4) -- (9);
\draw[line width=0.7] (9) -- (2);
\draw[line width=0.7] (2) -- (11);
\draw[line width=0.7] (0) -- (10);
\draw[line width=0.7] (4) -- (13);
\draw[line width=0.7] (6) -- (12);
\draw (-1.55,14.25) node{${\scriptstyle \bigomega_1}$};
\draw (17.6,14.25) node{${\scriptstyle \bigomega_2}$};
\draw (17.8,-2.25) node{${\scriptstyle \bigomega_3}$};
\draw (-1.55,-2.28) node{${\scriptstyle \bigomega_4}$};
\draw (2.55,12.75) node{${\scriptstyle \bigomega_{1}^{(1)}}$};
\draw (3.55,10.9) node{${\scriptstyle \bigomega_{1}^{(2)}}$};
\draw (-1.35,9.8) node{${\scriptstyle \bigomega_{1}^{(3)}}$};
\draw (13.5,12.75) node{${\scriptstyle \bigomega_{2}^{(1)}}$};
\draw (13.15,10.9) node{${\scriptstyle \bigomega_{2}^{(2)}}$};
\draw (17.35,9.8) node{${\scriptstyle \bigomega_{2}^{(3)}}$};
  \end{tikzpicture}
    \caption{\label{fig:Dualfactors}
The dual NFG of the Gaussian graphical model in Fig.~\ref{fig:Primalfactors} with PDF as in~(\ref{eqn:ModelPDFT2}). 
The boxes labelled ``$=$" represent equality indicator factors, the small empty boxes represent unary factors 
$\{(\mathcal{F}g_i)(\omega_i), i \in [n]\}$, and the big empty boxes represent pairwise 
factors $\{(\mathcal{F}h)(\omega_{i}, \omega_{j}), (i,j) \in \E\}$.}
  \end{figure}

From (\ref{eqn:ftdeltaplus}), the Fourier transform of the zero-sum indicator factor $\delta_{+}(\x_{\text{-}i}, y_i)$ is
\begin{equation} 
\label{eqn:deltatransG1}
 (2\pi)^{n-1} \delta_{=}(\omega_i, \omega_i^{(1)}, \omega_i^{(2)}, \ldots, \omega_i^{(n-1)})
\end{equation}
which enforces all the variables $\omega_i^{(1)}, \omega_i^{(2)}, \ldots, \omega_i^{(n-1)}$ to 
be equal to $\omega_i$. Put differently, in every valid configuration it holds that 
$\omega_i = \omega_i^{(1)} = \ldots = \omega_i^{(n-1)}$ as illustrated in Fig.~\ref{fig:Dualfactors}. 

\begin{figure}[t]
  \centering
    \begin{tikzpicture}[scale=0.27]
   \linethickness{1.7mm}
%
%
%
%
%

\node[draw, line width=0.7, circle, fill = black, label=below:{${\scriptstyle \bigomega_4}$}] at (0,0) (0) {$$};
\node[draw, line width=0.7, rectangle] at (8,0) (1) {\textcolor{white}{{\Large A}}};
\node[draw, line width=0.7, circle, fill = black, label=below:{${\scriptstyle \bigomega_3}$}] at (16,0) (2) {$$};
\node[draw, line width=0.7, rectangle] at (0,6) (3) {\textcolor{white}{{\Large A}}};
\node[draw, line width=0.7, circle, fill = black,, label=above:{${\scriptstyle \bigomega_1}$}] at (0,12) (4) {$$};
\node[draw, line width=0.7, rectangle] at (8,12) (5) {\textcolor{white}{{\Large A}}};
\node[draw, line width=0.7, circle, fill = black, label=above:{${\scriptstyle \bigomega_2}$}] at (16,12) (6) {$$};
\node[draw, line width=0.7, rectangle] at (16,6) (7) {\textcolor{white}{{\Large A}}};
\node[draw, line width=0.7, rotate=-52, rectangle] at (10.5,7.9) (8) {\textcolor{white}{{\Large A}}};
\node[draw, line width=0.7, rotate=52, rectangle] at (5.4,7.9) (9) {\textcolor{white}{{\Large A}}};
\node[draw, line width=0.7, rectangle] at (-3.75,-2.75) (10) {\textcolor{white}{{\tiny A}}};
\node[draw, line width=0.7, rectangle] at (19.75,-2.75) (11) {\textcolor{white}{{\tiny A}}};
\node[draw, line width=0.7, rectangle] at (19.75,14.75) (12) {\textcolor{white}{{\tiny A}}};
\node[draw, line width=0.7, rectangle] at (-3.75,14.75) (13) {\textcolor{white}{{\tiny A}}};
\draw[line width=0.7] (0) -- (1);
\draw[line width=0.7] (1) -- (2);
\draw[line width=0.7] (0) -- (3);
\draw[line width=0.7] (3) -- (4);
\draw[line width=0.7] (4) -- (5);
\draw[line width=0.7] (5) -- (6);
\draw[line width=0.7] (6) -- (7);
\draw[line width=0.7] (7) -- (2);
\draw[line width=0.7] (0) -- (8);
\draw[line width=0.7] (8) -- (6);
\draw[line width=0.7] (4) -- (9);
\draw[line width=0.7] (9) -- (2);
\draw[line width=0.7] (2) -- (11);
\draw[line width=0.7] (0) -- (10);
\draw[line width=0.7] (4) -- (13);
\draw[line width=0.7] (6) -- (12);
  \end{tikzpicture}
  \vspace{1.0ex}
  \caption{\label{fig:DualfactorsF}
The dual factor graph of the Gaussian graphical model in Fig.~\ref{fig:Primalfactors}, in which the vertices depicted by filled 
circles represent $\{\omega_i, i \in [n]\}$.}
\end{figure}
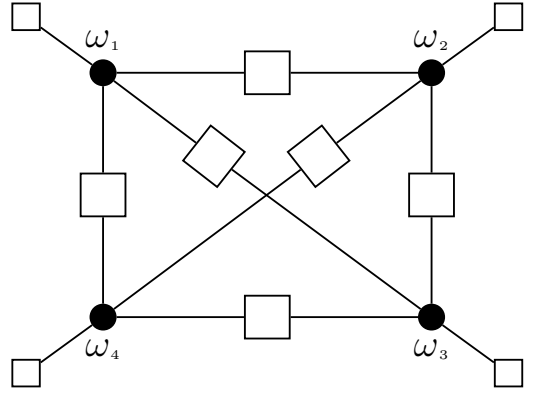

It is informative to draw the corresponding dual factor graph of the model, illustrated 
in Fig.~\ref{fig:DualfactorsF}. In a factor graph the variables are represented by vertices~\cite{KFL:01}, which are shown by filled black circles.  
It is clear from Fig.~\ref{fig:Dualfactors} that the transformed model contains $n$ independent 
variables $\{\omega_i, i \in [n]\}$.

The Fourier transform of the local factors~(\ref{eqn:defunaryg}) and~(\ref{eqn:defpairwiseh}) can be obtained from (\ref{eqn:ftGaussian}), which, for $ i \in [n]$, gives
\begin{equation}
\label{eqn:transS}
(\mathcal{F}g_i)(\omega_i) = (2\pi s_i^2)^{1/2}\textrm{exp}\Big(-\frac{s^2_i}{2}\omega^2_i\Big) 
\end{equation}
and for $ (i, j) \in \E$ gives
\begin{multline}
\label{eqn:transH}
(\mathcal{F}h)(\omega_{ij}, \omega_{ji}) = \\ \text{det}(2\pi\mathbf{K})^{1/2} \textrm{exp}\Big(-\frac{1}{2} 
\setlength\arraycolsep{2pt}
\begin{bmatrix}
\omega_{ij} & \omega_{ji}
\end{bmatrix}
\mathbf{K}
\begin{bmatrix}
\omega_{ij} \\
\omega_{ji}
\end{bmatrix}
\Big)
\end{multline}
with
\begin{equation}
\label{eqn:KInv}
\mathbf{K}  = \left[ \begin{array}{cc}
  \sigma^2 & \rho\sigma^2  \\[0.4ex]
  \rho\sigma^2 & \sigma^2
\end{array} \right]
\end{equation}
which is the inverse of the local information matrix in (\ref{eqn:K}).
 
For $n = 4$, the dual NFG (i.e., the dual of Fig.~\ref{fig:Primalfactors}) is shown in 
Fig.~\ref{fig:Dualfactors}, where
the boxes labelled ``$=$" represent equality indicator factors, the small empty boxes are unary factors 
$\{(\mathcal{F}g_i)(\omega_i), i \in [n]\}$, and the big empty boxes represent pairwise 
factors $\{(\mathcal{F}h)(\omega_{i}, \omega_{j}), (i,j) \in \E\}$.


The PDF associated with the dual model is thus
\begin{IEEEeqnarray}{lCl}
f(\pmb{\omega}) \propto
&\prod_{i \in \left[n\right]}& \delta_{=}(\omega_i, \omega_i^{(1)}, \ldots, \omega_i^{(n-1)}) \notag \\
&\prod_{i \in \left[n\right]}&(\mathcal{F}g_i)(\omega_i)\prod_{(i,j) \in \E} (\mathcal{F}h)(\omega_i, \omega_j) \label{eqn:ModelPDFT1}
\end{IEEEeqnarray}

which gives
\begin{multline}
f(\pmb{\omega}) =  \frac{1}{Z_{\pmb{\omega}}}\\[0.25ex]
 \prod_{i \in \left[n\right]}\textrm{exp}\left(-\frac{s^2_{i}}{2}\omega^2_i\right)
\prod_{(i,j) \in \E} \textrm{exp}\left(-\frac{1}{2} 
\setlength\arraycolsep{2pt}
\begin{bmatrix}
\omega_i & \omega_j
\end{bmatrix}
\mathbf{K}
\begin{bmatrix}
\omega_i \\
\omega_j
\end{bmatrix}
\right) 
\label{eqn:ModelPDFT2}
\end{multline}
where $Z_{\pmb{\omega}}$ is the appropriate normalization constant.

Since the quadratic form in~(\ref{eqn:ModelPDFT2}) only depends on $\omega_i^2$ and on $\omega_i\omega_j$, sign-inversion 
(step III in the dualization procedure) has no bearing on the resulting PDF. See~\cite{Mo:ITW2021} for a similar approach in the context 
of Gaussian graphical models defined on the ladder graph.

After collecting all the local scale factors in (\ref{eqn:deltatransG1}), (\ref{eqn:transS}), and (\ref{eqn:transH})  we obtain
\begin{equation*}
Z_{\pmb{\omega}} = (2\pi)^{N}\text{det}(2\pi\pmb{\Sigma}_{\pmb{\omega}})^{1/2}\text{det}(2\pi\mathbf{K})^{|\E|/2} \prod_{i \in \left[n\right]} (2\pi s_i^2)^{1/2} 
\end{equation*}
Hence
\begin{equation}
Z_{\pmb{\omega}} = (2\pi)^{3N/2}\text{det}(\pmb{\Sigma}_{\pmb{\omega}})^{1/2}\text{det}(\mathbf{K})^{|\E|/2} \prod_{i \in \left[n\right]} s_i \label{eqn:ZWForm}
\end{equation}

Here $\pmb{\Sigma}_{\pmb{\omega}} \in \R^{n\times n}$ is the information matrix associated with $f(\pmb{\omega})$ in (\ref{eqn:ModelPDFT2}).
The entry in the $i$-th row and the $j$-th column of  $\pmb{\Sigma}^{-1}_{\pmb{\omega}}$ is given by
\begin{equation} 
\label{eqn:SigmaDual}
\pmb{\Sigma}^{-1}_{\pmb{\omega}}[i,j] = \left\{ \begin{array}{ll}
       s_i^2 + (n-1)\sigma^2, & \text{if $i = j$,} \\[1.1ex]
      \rho\sigma^2, & \text{otherwise.} 
      \end{array}\right.   
\end{equation}

The information matrix $\pmb{\Sigma}^{-1}_{\pmb{\omega}}$ can be decomposed as 
\begin{equation}
\label{eqn:DecDual}
\pmb{\Sigma}^{-1}_{\pmb{\omega}}  = \mathbf{D} + \rho\sigma^2 \mathbf{J}_n
\end{equation}
where $\mathbf{D} \in \R^{n\times n}$ is a diagonal matrix given by 
\begin{equation} 
\label{eqn:K2}
\mathbf{D}[i,j] = \left\{ \begin{array}{ll}
       s_i^2 + (n - \rho - 1)\sigma^2, & \text{if $i = j$,} \\[1.1ex]
       0, & \text{otherwise.} 
      \end{array}\right.   
\end{equation}

According to the NFG Duality Theorem, $Z_{\pmb{\omega}}$ and $Z_\x$ are equal up to scale. Indeed
\begin{equation}
\label{eqn:NFGScale}
Z_{\pmb{\omega}} = (2\pi)^{N}Z_\x
\end{equation}
where $N$ is the total number of variables in (\ref{eqn:ModelPDF2}) given by (\ref{eqn:DefN}). 

After plugging (\ref{eqn:Zx2}) and (\ref{eqn:ZWForm}) 
into (\ref{eqn:NFGScale}), we obtain
\begin{equation}
\label{eqn:SigmaXtoW}
\text{det}(\pmb{\Sigma}_{\pmb{\omega}})\,\text{det}(\mathbf{K})^{|\E|} \prod_{i \in \left[n\right]} s^2_i = \text{det}(\pmb{\Sigma}_{\x})
\end{equation}
or equivalently 
\begin{equation}
\label{eqn:SigmaXtoWAgain}
\text{det}(\pmb{\Sigma}_{\pmb{\omega}}^{-1}) = \text{det}(\pmb{\Sigma}_{\x}^{-1})\,\text{det}(\mathbf{K})^{|\E|} \prod_{i \in \left[n\right]} s^2_i 
\end{equation}

We will apply the Matrix Determinant Lemma in our proof, which states that if
$\mathbf{B} \in \R^{p\times p}$ is an invertible matrix and $\mathbf{u}, \mathbf{v} \in \R^p$ are two column vectors, then
\begin{equation}
\label{eqn:MatrixDeterminantLemma}
\text{det}(\mathbf{B} + \mathbf{uv}^\textsf{T}) = (1 + \mathbf{v}^\textsf{T}\mathbf{B}^{-1}\mathbf{u})\text{det}(\mathbf{B})
\end{equation}
For more details, see Sherman-Morrison-Woodbury formula in~\cite{horn2013matrix}.

{\bf Proposition.} The closed-form expression for $\text{det}(\pmb{\Sigma}_\x)$ is 
\begin{equation}
\label{eqn:DetResult}
\text{det}(\pmb{\Sigma}_\x) = 
\frac{\text{det}(\mathbf{K})^{|\E|}\prod_{i \in \left[n\right]} s^2_i}{\big(1+ \rho\sigma^2\text{tr$(\mathbf{D}_{}^{-1})$}\big)\text{det}(\mathbf{D})}
\end{equation}

\emph{Proof.} Starting from (\ref{eqn:DecDual})
\begin{align}
\text{det}(\pmb{\Sigma}^{-1}_{\pmb{\omega}}) & = \text{det}(\mathbf{D} + \rho\sigma^2\mathbf{J}_n)\\[0.25ex]
               & = \big(1+ \rho\sigma^2\sum_{i \in \left[n\right]}\mathbf{D}_{}^{-1}[i,i]\big)\text{det}(\mathbf{D}) \label{eqn:UseDetLema}\\
               & = \big(1+ \rho\sigma^2\text{tr$(\mathbf{D}_{}^{-1})$}\big)\text{det}(\mathbf{D}) \label{eqn:UseDetLema2}
\end{align}

Here (\ref{eqn:UseDetLema}) follows form the Matrix Determinant Lemma (\ref{eqn:MatrixDeterminantLemma}) by setting
$\mathbf{u} = \mathbf{v} = \rho^{1/2}\sigma\mathbf{1}_n$.

The Proposition follows from plugging 
in $\text{det}(\pmb{\Sigma}_{\pmb{\omega}})$ from (\ref{eqn:SigmaXtoWAgain}) in~(\ref{eqn:UseDetLema2}).


\hfill$\Box$


Let $s_i = s$ for $i \in [n]$ and $d = s^2 + (n - \rho - 1)\sigma^2$. Therefore $\mathbf{D} = d\mathbf{I}_n$ 
from (\ref{eqn:K2}) and $\pmb{\Sigma}^{-1}_{\pmb{\omega}}  = d\mathbf{I}_n + \rho\sigma^2 \mathbf{J}_n$. The closed-form solution of this homogeneous model
can be expressed as~\footnote{In a homogeneous model $\pmb{\Sigma}^{-1}_{\pmb{\omega}}$ is closely related to the Pei matrix, which 
has the form $\mu\mathbf{I}_n + \mathbf{1}_n^{}\mathbf{1}_n^\textsf{T}$. The Pei matrix is symmetric, and is positive 
definite if $\mu > 0$. For more details see~\cite{higham1995test}.}
\begin{equation}
\label{eqn:DetResultSp}
\text{det}(\pmb{\Sigma}_{\x})  = \frac{\text{det}(\mathbf{K})^{|\E|}s^{2n}}{(1+n\rho\sigma^2/d)d^n}
\end{equation}
In the limit $N \to \infty$, we obtain
\begin{align}
\lim_{N \to \infty}\frac{\ln\text{det}(\pmb{\Sigma}_{\x})}{N} & = \lim_{N \to \infty}\frac{|\E|}{N}\ln\text{det}(\mathbf{K}) \\ 
& = 2\ln\sigma + \frac{1}{2}\ln \left(1-\rho^2\right) \label{eqn:DetResultLim2}
\end{align}
Recall from Section~\ref{sec:notations} that $N = 2|\E| = n(n-1)$ and form (\ref{eqn:KInv}), we have $\text{det}(\mathbf{K}) = \sigma^4(1-\rho^2)$.

According to (\ref{eqn:SigmaXtoW}) the determinants of the covariance matrix of the original (primal) model
and the covariance matrix of its dual model are equal up to scale. Indeed
\begin{equation}
\label{eqn:EquivAgain}
\frac{\text{det}(\pmb{\Sigma}_{\x})}{\text{det}(\pmb{\Sigma}_{\pmb{\omega}})} = \text{det}(\mathbf{K})^{|\E|} \prod_{i \in \left[n\right]} s^2_i
\end{equation}
In this context, we may call the Gaussian graphical models in Figs.~\ref{fig:Primalfactors} and \ref{fig:Dualfactors} \emph{equivalent}, as their normalization constants
are equal up to a scale factor (which is easy to compute).

%
%

\section{Numerical examples}
\label{sec:ExampleT}

In our first example, we again look at the model described in Section~\ref{sec:Example}.
From (\ref{eqn:DecDual}) we can write
\begin{equation}
\pmb{\Sigma}^{-1}_{\pmb{\omega}} =\mathbf{D}  + \frac{1}{3}\mathbf{J}_4
\end{equation}
where $\mathbf{D} = \text{diag$(8/3, 13/6, 2, 23/12)$}$. After applying the 
Proposition, it is easy to compute $\ln\text{det}(\pmb{\Sigma}_{\x}) \approx -13.35$.

We consider $\calK_n$ in the second example. We assume that the model is homogenous, and set $\rho = -4/5$, $\sigma^2 = 1$, and $s^2 = 1$. 
E.g., for $N=6$ (i.e., $n=3$), the information matrix of the model is given by
\begin{equation}
\pmb{\Sigma}^{-1}_{\x}  = \frac{1}{9}
\left[ \begin{array}{cccccc}
 34 & 9 & 20 & 0 & 0 & 0 \\
 9 & 34 & 0 & 0 & 20 & 0 \\
 20 & 0 & 34 & 9 & 0 & 0 \\
 0 & 0 & 9 & 34 & 0 & 20 \\
 0 & 20 & 0 & 0 & 34 & 9 \\
 0 & 0 & 0 & 20 & 9 & 34
\end{array} \right]
\end{equation}
For this size of the model $\mathbf{D} = 19/5\mathbf{I}_{3}$, and
\begin{equation}
\pmb{\Sigma}^{-1}_{\pmb{\omega}}  = \frac{1}{5}
\left[ \begin{array}{rrr}
 15 & -4 & -4 \\
 -4 & 15 & -4  \\
 -4 & -4 & 15 
\end{array} \right]
\end{equation}

From (\ref{eqn:DetResultSp}), we can compute the exact $\text{det}(\pmb{\Sigma}_{\x})$, which is
\begin{equation}
\text{det}(\pmb{\Sigma}_{\x}) = 
\frac{3^{n(n-1)}}{(n+4)(5n+4)^{n-1}5^{n(n-2)}}
\end{equation}
for an arbitrary $n$. In the limit $N \to \infty$
\begin{equation}
\lim_{N \to \infty}\frac{\ln\text{det}(\pmb{\Sigma}_{\x})}{N} = \ln\big(\frac{3}{5}\big)
\end{equation}
which can also be obtained directly from (\ref{eqn:DetResultLim2}).

\section{Conclusion}
\label{sec:Conclusion}

We provided a closed-form 
solution for the determinant of the covariance matrix of a specific class of sparse Gaussian graphical models whose covariance selection model is equal to $\calK_{n} \ccirc\, \calK_{n-1}$, where $\calK_n$ is the complete graph with $n$ vertices 
and $\ccirc$ stands for the replacement product. The closed-form solution was obtained by a straightforward 
application of the Matrix Determinant Lemma applied on the dual NFG of the
model. 

\balance

\bibliography{mybib}

\end{document}